\documentclass{ieeeaccess}
\usepackage{cite}
\usepackage{amsmath,amssymb,amsfonts}
\usepackage{algorithmic}
\usepackage{graphicx}
\usepackage{textcomp}

\usepackage{bm}
\usepackage{booktabs}
\usepackage{array}
\usepackage{algorithm}
\usepackage{algorithmic}

\newcolumntype{L}[1]{>{\raggedright\let\newline\\\arraybackslash\hspace{0pt}}m{#1}}
\newcolumntype{C}[1]{>{\centering\let\newline\\\arraybackslash\hspace{0pt}}m{#1}}
\newcolumntype{R}[1]{>{\raggedleft\let\newline\\\arraybackslash\hspace{0pt}}m{#1}}

\def\BibTeX{{\rm B\kern-.05em{\sc i\kern-.025em b}\kern-.08em
    T\kern-.1667em\lower.7ex\hbox{E}\kern-.125emX}}
\begin{document}
\history{Date of publication xxxx 00, 0000, date of current version xxxx 00, 0000.}
\doi{10.1109/ACCESS.2017.DOI}

\title{Collaborative Training of GANs in Continuous and Discrete Spaces for Text Generation}
\author{Yanghoon Kim\authorrefmark{1},
Seungpil Won\authorrefmark{1}, Seunghyun Yoon\authorrefmark{2} and Kyomin Jung\authorrefmark{1}
}
\address[1]{Seoul National University, Seoul, Korea}
\address[2]{Adobe Research, San Jose, CA, USA}

\markboth
{Author \headeretal: Preparation of Papers for IEEE TRANSACTIONS and JOURNALS}
{Author \headeretal: Preparation of Papers for IEEE TRANSACTIONS and JOURNALS}

\corresp{Corresponding author: Kyomin Jung (e-mail: kjung@snu.ac.kr).}

\begin{abstract}
Applying generative adversarial networks (GANs) to text-related tasks is challenging due to the discrete nature of language.
One line of research resolves this issue by employing reinforcement learning (RL) and optimizing the next-word sampling policy directly in a discrete action space.
Such methods compute the rewards from complete sentences and avoid error accumulation due to exposure bias.
Other approaches employ approximation techniques that map the text to continuous representation in order to circumvent the non-differentiable discrete process.
Particularly, autoencoder-based methods effectively produce robust representations that can model complex discrete structures.
In this paper, we propose a novel text GAN architecture that promotes the collaborative training of the continuous-space and discrete-space methods.
Our method employs an autoencoder to learn an implicit data manifold, providing a learning objective for adversarial training in a continuous space.
Furthermore, the complete textual output is directly evaluated and updated via RL in a discrete space.
The collaborative interplay between the two adversarial trainings effectively regularize the text representations in different spaces.
The experimental results on three standard benchmark datasets show that our model substantially outperforms state-of-the-art text GANs with respect to quality, diversity, and global consistency.
\end{abstract}


\titlepgskip=-15pt

\maketitle

\section{Introduction}
\label{sec:introduction}
Generating realistic text is an important task with a wide range of real-world applications, such as machine translation~\cite{wu2016google}, dialogue generation \cite{li2016deep}, image captioning \cite{vinyals2015caption}, and summarization \cite{allahyari2017text}.
A language model is the most common approach for text generation, and it is typically trained via maximum likelihood estimation (MLE), specifically in an autoregressive fashion.

Although MLE-based methods have achieved great success in text generation, there are two fundamental issues that call for further research. The first problem is that MLE suffers from \textit{exposure bias} \cite{bengio2015scheduled}: during training, the model sequentially generates the next word depending on the ground-truth words; however, the model relies on its previously generated words at inference time.
Therefore, the cumulative effect of incorrect predictions in the text sequence results in low-quality samples.
The second problem lies in that the objective functions of MLE-based methods are rigorous~\cite{press2017language}; the models are forced to learn every word in the target sentence. Under this strict guidance, the ability of language models to generate diverse samples can be severely limited.

In recent years, Generative Adversarial Networks (GANs) \cite{goodfellow2014gan} have drawn attention as a remedy to the above problems. However, applying GANs to text-related tasks is challenging due to the discrete nature of text. In the inference phase of the text generation, the model iteratively samples the next word from the distribution of vocabulary. As this step includes the sampling process that hinders the backpropagation of the gradients from the discriminator, several approximation methods have been proposed to avoid the non-differentiability issue~\cite{yu2017seqgan, lin2017adversarial, zhao2018adversarially, subramanian2018towards}.

Depending on the data space in which the GAN's discriminator operates, text GANs can be categorized into two groups: continuous-space methods and discrete-space methods.  
For discrete spaces, one prominent research line adopts the reinforcement learning (RL) technique to address the non-differentiability issue directly~\cite{yu2017seqgan, lin2017adversarial, de2019training}. In the RL setting, GANs treat the generator as a stochastic policy to synthesize realistic samples. The generator is optimized via policy gradient methods by incorporating the reward signals from the discriminator. These signals are computed from a complete sequence rather than individual words in text.
This RL approach can consider the final form of the text, thus it resolves the discrepancy between the training and inference stages in the MLE method. 
However, it has significant limitations, such as an excessive dependency on MLE pretraining, and severe mode collapse~\cite{lu2019cot}.

Other methods employ approximation techniques to transform discrete text into continuous representation. Such approaches include substituting next-word sampling in the generation phase with continuous relaxation~\cite{kusner2016gans,zhang2017adversarial} or adopting an autoencoder architecture to learn an implicit data manifold in a continuous space instead of directly modeling the discrete text~\cite{zhao2018adversarially, haidar2019latent}.
In these approaches, the discriminator distinguishes between the synthetic and real text representations in the continuous space.
As the discriminator only learns to distinguish an approximated representation of text, these approaches cannot provide direct feedback concerning the entire text's correctness.

In this work, we propose a novel text GAN architecture, called ConcreteGAN, which promotes the collaborative training of the \underline{\textbf{\textit{cont}}}inuous-space and dis\underline{\textbf{\textit{crete}}}-space methods.
Specifically, in the continuous space, a latent code representation of the synthetic text is learned jointly with an autoencoder. 
Then, the textual output generated from the latent code is further updated via RL training.
In this way, ConcreteGAN simultaneously regularizes the text generation process within the continuous and discrete data spaces.
The interplay between adversarial trainings in the two data spaces takes the following advantages; 1) it reduces RL training variance through the regularization of the latent code representation; and 2) it alleviates exposure bias in continuous-space methods.
To the best of our knowledge, our proposed method is the first work to train a text GAN combining both continuous-space and discrete-space methods.

We evaluate our model on three benchmark datasets: the COCO Image Caption corpus, the Stanford Natural Language Inference corpus, and the EMNLP 2017 WMT News corpus. Extensive experiments show that our model surpasses the existing text GAN models and achieves a substantial improvement with respect to quality, diversity, and global consistency.
In addition, we provide comprehensive analyses of the latent code space. Compared to the GANs that work only in a continuous space, the synthetic code space generated by our model is more similar to the latent code space of real text.
This behavior demonstrates that the proposed approach effectively regularizes the latent code space, which helps to reduce the variance of RL training.

\section{Background}
In this section, we first give a brief description of GANs. Then we introduce two lines of research for text generation, including continuous-space methods and discrete-space methods.

\subsection{Generative Adversarial Networks}
GANs are one of the implicit generative models that do not require a tractable likelihood function. Thus, they can be applied to practical situations such as imitating the distribution of high-dimensional complex data.

In general, GANs have a generator and a discriminator as their basic components. The generator tries to imitate the real data distribution, and the discriminator tries to distinguish generated samples from real data. The iterative interplay between these two components improves their strength against each other and provides significant performance enhancement in each of them. One can formulate the objective function of the GAN as a minimax game:

\begin{equation}
\begin{aligned}
    \underset{G}{\mathrm{min}}\,\underset{D}{\mathrm{max}} \; F(D,G) &= \mathbb{E}_{\mathbf{x} \sim data}[\log D(\mathbf{x})] \\
    &+ \mathbb{E}_{\mathbf{x} \sim noise}[\log (1 - D(G(\mathbf{x})))], \\
\end{aligned}
\end{equation}
where $G$ and $D$ are the functions for the generator and discriminator, respectively.

GANs have shown significant achievements in various deep learning applications, especially in computer vision research~\cite{radford2015unsupervised, zhu2017unpaired, karras2019style}. However, when applied to text generation, GANs suffer from the non-differentiability issue due to the discrete nature of text.
Recently, various methods have been proposed to circumvent this issue, which can be broadly classified into two categories: continuous-space methods and discrete-space methods.

\Figure(topskip=0pt, botskip=0pt, midskip=0pt)[width=\linewidth]{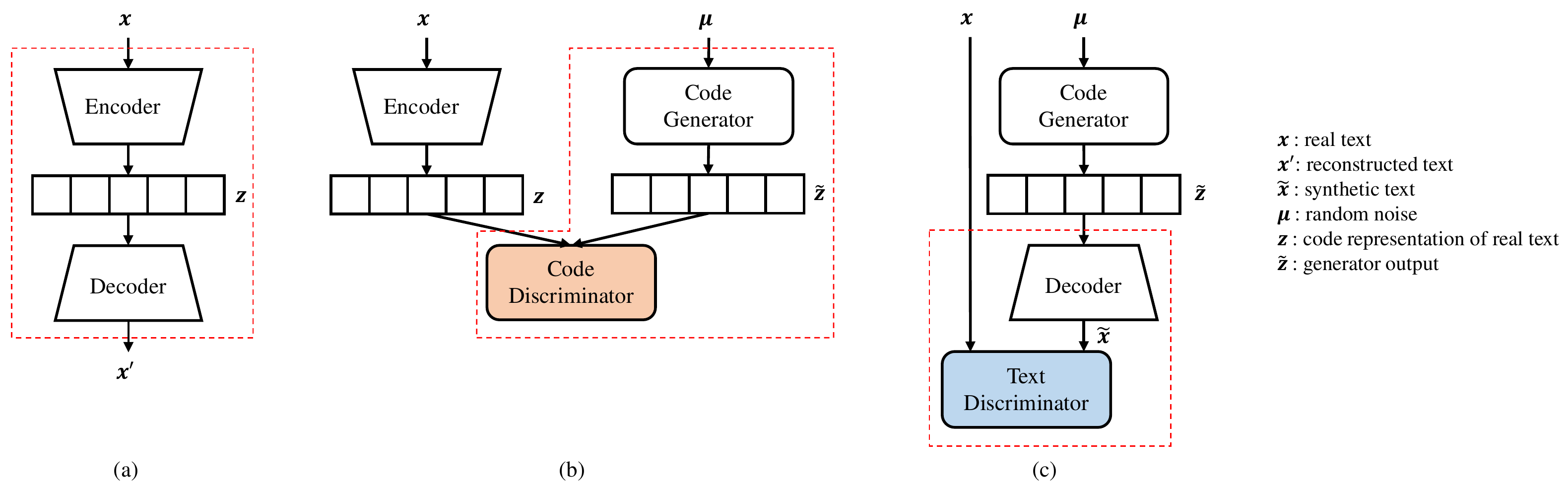}
{The overall architecture of our model, which is composed of an autoencoder, a code-generator, a code-discriminator and a text-discriminator. Each training iteration of our model has three steps: (a) autoencoder reconstruction, (b) adversarial training of code-generator and code-discriminator in a continuous space, and (c) adversarial training of decoder and text-discriminator in a discrete space. 
The red dotted lines specify the modules to be updated in each step. The collaborative interplay between two adversarial training in continuous (latent code) and discrete (natural text) space regularizes the text representations in different spaces.}
\label{fig_model}

\subsection{Continuous-space Methods}
Several studies on GANs sidestep the non-differentiability by reformulating the learning objective in a continuous space. The adversarially regularized autoencoder (ARAE)~\cite{zhao2018adversarially}, as a representative model, employs an autoencoder to learn an implicit data manifold, mapping discrete text into a continuous latent representation. In this model, the encoder and the generator are trained adversarially. Based on the ARAE, LATEXT-GAN~\cite{haidar2019latent} uses an additional approximated representation of text called soft-text, which is the reconstructed output of the autoencoder. Then, they employs two discriminators for each approximated representation in the continuous spaces.

Another line of work exploits a differentiable continuous relaxation, (i.e., Gumbel-softmax~\cite{kusner2016gans, nie2018relgan} or soft-argmax~\cite{zhang2017adversarial}), to replace next-word sampling in the generation phase.

\subsection{Discrete-space Methods}
SeqGAN~\cite{yu2017seqgan} is the first work addressing the non-differentiablity issue within a discrete space by introducing an RL technique into GAN training. Specifically, this approach considers the generated words as the current state and the generation of the next word as an action. In this scenario, the generator is optimized via a policy gradient, where the reward is computed by the discriminator through a Monte Carlo search. MaliGAN~\cite{che2017maximum} proposes a normalized maximum likelihood objective. Combined with several reduction techniques, it reduces the variance of the reinforcement learning rewards and the instability of the GAN training dynamics. LeakGAN~\cite{guo2018long} devises a hierarchical architecture for the generator to address the sparsity issue in the long text generation. The generator is guided by the latent feature leaked from the discriminator at all generation steps. RankGAN~\cite {lin2017adversarial} relaxes the binary restriction of the discriminator by exploiting relative ranking information between the real sentences and generated ones. This increases the diversity and richness of the sentences. All of the above methods use the maximum likelihood pretraining, followed by small amounts of adversarial fine-tuning. ScratchGAN~\cite{de2019training} first achieved a performance comparable to that of MLE methods without any pretraining.

\section{ConcreteGAN}
In this work, we propose a novel text GAN architecture that promotes the collaboration of two adversarial trainings in a continuous space and a discrete space, respectively. We adopt the alternating training of the two methods in each iteration, rather than MLE pretraining commonly used for discrete-space methods. The architecture of our model is shown in Fig. 1.

Our model consists of the following four components: (1) RNN-based autoencoder is composed of an encoder and a decoder. The encoder provides a latent code representation of real text in a continuous space. The decoder, as a text-generator, yields textual outputs by interpreting the latent code from encoder or code-generator. (2) Code-generator maps a random noise to a latent code representation with the goal to imitate the distribution of the encoder. (3) Code-discriminator is the code-generator's opponent, and is adversarially trained to distinguish between latent codes from encoder and code-generator. (4) Text-discriminator evaluates complete sequences from real data distribution or decoder(or text-generator) distribution. The computed scores are used as the rewards to train the decoder via the policy gradient algorithm.
The interplay between two adversarial trainings has a complementary effect, improving both the quality and the diversity of generated text.

\subsection{Autoencoder Reconstruction}
Let $\mathbf{x}\in\mathbf{X}$ be the input sequence and $\mathbf{z}\in\mathbf{Z}$ be the latent code of an autoencoder. We use a conventional RNN autoencoder that consists of two parts: an encoder network and a decoder network. The encoder network $f_{\phi} : \mathbf{X}\mapsto\mathbf{Z}$ (parameterized by $\phi$) maps the input sequence $\mathbf{x}$ to the latent code $\mathbf{z}$, which is represented as the last hidden state. The decoder function $f_{\psi} : \mathbf{Z}\mapsto\mathbf{X}$ (parameterized by $\psi$) reconstructs the original input $\mathbf{x}$ conditioned on the encoded latent code $\mathbf{z}$. Here, we use a gated recurrent unit (GRU) for both the encoder and decoder networks, whose parameters are trained using the cross-entropy loss function:

\begin{gather}
    \mathbf{z} = f_{\phi }(\mathbf{x}) \label{eq_2} \\
    \underset{\phi, \psi}{\mathrm{min}} \; \mathit{L}_{rec}(\phi, \psi) = - \mathbb{E}[\log p_\psi(\mathbf{x} | \mathbf{z})] \label{eq_3}
\end{gather}

\subsection{Adversarial Training in the Latent Code Space}
The next step of autoencoder reconstruction is the adversarial training of code-generator $G_{\theta}(\bm{\mu})$ and code-discriminator $D_{\omega}(\mathbf{z})$. The code-generator aims to imitate the distribution of real text in the continuous latent code space that is represented as the last hidden state of the encoder. Given a random noise vector $\bm{\mu}$ from a fixed distribution, such as a standard Gaussian distribution, the code-generator $G_{\theta}(\bm{\mu})$ outputs a vector $\mathbf{\tilde{z}}$ that has the same shape as the last hidden state of the encoder.
On the other hand, the code-discriminator $D_{\omega}(\mathbf{z})$ learns to distinguish code-generator's output from the latent representations of real text. We use multilayer perceptrons (MLPs) with residual connections for both $G_{\theta}(\bm{\mu})$ and $D_{\omega}(\mathbf{z})$ and adopt WGAN with gradient penalty (WGAN-GP) for optimization.
    
\begin{gather}
    \mathbf{\tilde{z}} = G_{\theta}(\bm{\mu}) \\
    \begin{split}
        \underset{\omega}{\mathrm{min}} \; L_{\omega} &= \underset{\mathbf{\tilde{z}} \sim \mathbb{P}_G}{\mathbb{E}} [D_{\omega}(\mathbf{\tilde{z}})] - \underset{\mathbf{z} \sim \mathbb{P}_{f_\phi}}{\mathbb{E}} [D_{\omega}(\mathbf{z})] \\
        &+ \lambda \underset{\mathbf{\hat{z}} \sim \mathbb{P}_{\mathbf{\hat{z}}}}{\mathbb{E}} [(\left \| \nabla_{\mathbf{\hat{z}}} D_{\omega}(\mathbf{\hat{z}}) \right \|_2 - 1)^2]
    \end{split} \\
    \underset{\theta}{\mathrm{min}} \; L_{\theta} = - \underset{\mathbf{\tilde{z}} \sim \mathbb{P}_G}{\mathbb{E}} [D_{\omega}(\mathbf{\tilde{z}})]
\end{gather}
where $\mathbf{\hat{z}} = t \mathbf{\tilde{z}} + (1-t) \mathbf{z}$ with $0 \leq t \leq 1$

\begin{table}[t]
\caption{Statistics of the standard benchmark datasets for evaluating text GANs.}
\small
\centering
\begin{tabular}{L{0.40\columnwidth}|C{0.125\columnwidth}C{0.125\columnwidth}C{0.135\columnwidth}}
\toprule
                 & COCO          & SNLI  & EMNLP  \\ \midrule
Vocabulary size         & 4,682          & 42,423  & 5,255 \\
Sequence length         & \textless 37  & \textless 81  & \textless 51 \\
\# of sentences (train) & 10k           & 701k  & 270k \\ 
\# of sentences (dev)   & 10k           & 13k  & 10k \\
\# of sentences (test)  & 10k           & 13k  & 10k \\ \bottomrule
\end{tabular}

\label{table_data}
\end{table}

\begin{table*}[t]
\caption{Comparison of the generated sentence quality and diversity in terms of the corpus-level BLEU and B-BLEU scores on the EMNLP 2017 WMT News dataset. A larger value indicates better quality / diversity.}
\small
\centering
\begin{tabular}{C{0.24\columnwidth}|C{0.20\columnwidth}C{0.20\columnwidth}C{0.20\columnwidth}C{0.20\columnwidth}C{0.20\columnwidth}C{0.20\columnwidth}C{0.20\columnwidth}}
\toprule
Metrics & MLE & SeqGan & RankGan & MaliGan & ScratchGan & ARAE* & Ours \\
\midrule
BLEU-2  & 0.843    & 0.761    & 0.736     & 0.764      & 0.830   & 0.824 & \textbf{0.858}\\
BLEU-3  & 0.573    & 0.463    & 0.441     & 0.468      & 0.552   & 0.614 & \textbf{0.658}\\
BLEU-4  & 0.328    & 0.228    & 0.204     & 0.231      & 0.309   & 0.391 & \textbf{0.431}\\
BLEU-5  & 0.182    & 0.115    & 0.095     & 0.113      & 0.172   & 0.228 & \textbf{0.257} \\
\midrule
B-BLEU-2  & \textbf{0.842}    & 0.693   & 0.671     & 0.684     & 0.832   & 0.808 & 0.806\\
B-BLEU-3  & \textbf{0.571}    & 0.413   & 0.373     & 0.391     & 0.559   & 0.552 & 0.555\\
B-BLEU-4  & 0.328    & 0.216   & 0.191     & 0.197     & 0.317   & 0.330 & \textbf{0.338} \\
B-BLEU-5  & 0.189    & 0.112   & 0.096     & 0.094     & 0.176   & 0.189 & \textbf{0.197} \\
\bottomrule 
\end{tabular}
\label{table_emnlp_bleu_bblue}
\end{table*}

\subsection{Adversarial Training with Textual Outputs}
Along with the adversarial training in the latent code space, we build another adversarial training loop that operates on the discrete textual outputs.

Given a code-generator with fixed weights, we model the decoder, which yields the textual outputs, as a policy and apply policy gradient method to optimize it. The text-discriminator $D_{\rho}$ is utilized to evaluate the generated sequence and provide the reward $R_t$.
Following previous works, we use REINFORCE \cite{williams1992simple}, a Monte Carlo (MC) variant of the policy gradient algorithm, for gradient estimation of the decoder training.

Since the reward signal can be calculated only when the entire sequence is completely generated, several approximation methods are proposed to obtain an intermediate reward for each generated token. While an MC search with a roll-out policy~\cite{yu2017seqgan} is the method adopted in most research, it is computationally expensive even with a feed-forward discriminator.
From our preliminary experiments, we find that GRU-based sequential discriminator shows better performance than a CNN discriminator with MC search in terms of computation time and evaluation results. With this empirical intuition, we use GRU-based discriminator as follows:
\begin{align}
    \begin{split}
        \underset{\rho}{\mathrm{min}} \; L_{\rho} &= -\sum_{t=1}^{T}\mathbb{E}_{\mathbf{x} \sim data}[\log D_{\rho}(x_t|x_1, ..., x_{t-1})] \\
        &+ \sum_{t=1}^{T'}\mathbb{E}_{\mathbf{\tilde{x}} \sim f_{\psi}(\tilde{z})}[\log D_{\rho}(\tilde{x}_t|\tilde{x}_1, ..., \tilde{x}_{t-1})]
    \end{split} \\
    R_t &= \sum_{s=t}^{T}\gamma^{s-t} D_{\rho}(\tilde{x}_t | \tilde{x}_{t-1}, ..., \tilde{x}_1) \\
    \nabla_{\psi} &= \sum_{t=1}^{T} \sum_{n=1}^{N} \nabla_{\psi} \log f_{\psi}(\tilde{x}_t|\tilde{x}_1, ..., \tilde{x}_{t-1}, \mathbf{z}) R_t
\end{align}
where $\gamma$ is a discount factor such that $0 < \gamma < 1$ and $N$ is the size of the mini-batch. The overall learning procedure is shown in Algorithm~\ref{main_algorithm}.


As a result of adversarial training in a continuous space, code-generator can provide an regularized latent representation of text sequence.
This leads to the effective restriction on the search space of the RL-policy decoder, acting as a guideline for generating a sentence within bounded space.
In adversarial training in a discrete space, the decoder learns to better capture the structure of text, such as a phrase, rather than the choice of words.
This process contributes to mitigate the exposure bias of autoencoder, which further affects the training process of continuous space.

\begin{algorithm}[t]
\caption{ConcreteGAN Training}
\label{main_algorithm}
\begin{algorithmic}
\REQUIRE text encoder $f_\phi$; shared decoder $f_\psi$; code-generator $G_{\theta}$; code-discriminator $D_{\omega}$; text-discriminator $D_{\rho}$; real text data $\mathbf{x}\in\mathbf{X}$; random noise vector $\bm{\mu}$;
\FOR{each training iteration}
    \STATE \textbf{(1) Train the autoencoder for reconstruction} ($f_{\phi}$, $f_{\psi}$) \\
    Compute $\mathbf{x'} = f_{\psi}(f_{\phi}(\mathbf{x}))$ \\
    Backprop loss $L_{rec}(\mathbf{x}, \mathbf{x'})$ \\
    
    \STATE \textbf{(2) Adversarial training in the code space} ($G_{\theta}$, $D_{\omega}$) \\
    Compute $\mathbf{z} = f_{\phi }(\mathbf{x})$ and $\mathbf{\tilde{z}} = G_{\theta}(\bm{\mu})$ \\
    Backprop loss $L_{\omega}(\mathbf{z}, \mathbf{\tilde{z}})$ to update $D_{\omega}$ \\
    Backprop loss $L_{\theta}(\mathbf{\tilde{z}})$ to update $G_{\theta}$ \\
    
    \STATE \textbf{(3) Adversarial training with textual output} ($f_{\psi}$, $D_{\rho}$) \\
    Compute $\mathbf{\tilde{x}} = f_{\psi}(G_{\theta}(\bm{\mu}))$ \\
    Backprop loss $L_{\rho}(\mathbf{x}, \mathbf{\tilde{x}})$ \\
    Train $f_{\psi}$ via policy gradient $\nabla_{\psi}$
    
\ENDFOR

\end{algorithmic}
\end{algorithm}

\begin{table*}[t]
\caption{Comparison of the generated sentence quality and diversity in terms of the corpus-level BLEU and B-BLEU scores on the COCO caption dataset. A larger value indicates the better quality / diversity.}
\small
\centering
\begin{tabular}{C{0.24\columnwidth}|C{0.20\columnwidth}C{0.20\columnwidth}C{0.20\columnwidth}C{0.20\columnwidth}C{0.20\columnwidth}C{0.20\columnwidth}C{0.20\columnwidth}}
\toprule
Metrics & MLE & SeqGan & RankGan & MaliGan & ScratchGan & ARAE* & Ours \\ 
\midrule
BLEU-2  & 0.745    & 0.748    & 0.727    & 0.733      & \textbf{0.762}   & 0.760 & 0.729\\
BLEU-3  & 0.509    & 0.514    & 0.491    & 0.497      & 0.535   & \textbf{0.557} & 0.526 \\
BLEU-4  & 0.317    & 0.307    & 0.291    & 0.295      & 0.344   & \textbf{0.374} & 0.347 \\
BLEU-5  & 0.196    & 0.187    & 0.175    & 0.178      & 0.221   & \textbf{0.243} & 0.222 \\
\midrule
B-BLEU-2  & \textbf{0.775}    & 0.748    & 0.727    & 0.733    & 0.762   & 0.766 & 0.756 \\
B-BLEU-3  & \textbf{0.552}    & 0.514    & 0.491    & 0.497    & 0.535   & 0.547 & 0.536 \\
B-BLEU-4  & 0.342    & 0.307    & 0.291    & 0.295    & 0.344   & \textbf{0.359} & 0.351 \\
B-BLEU-5  & 0.219    & 0.187    & 0.175    & 0.178    & 0.221   & \textbf{0.228} & 0.226\\ \bottomrule
\end{tabular}
\label{table_coco_bleu_bbleu}
\end{table*}

\section{Experiments}
To demonstrate the efficacy of our proposed method, we evaluate our model on various real-world datasets. In what follows, we give a detailed description of the whole evaluation process, from the experimental settings to the experimental results. We provide a performance comparison with state-of-the-art models as well as several analyses on the code space.

\subsection{Dataset}
We carry out experiments on three standard benchmark datasets for evaluating text GANs: COCO Image Caption (COCO) dataset \cite{chen2015microsoft}, Stanford Natural Language Inference (SNLI) corpus \cite{bowman2015snli} and EMNLP 2017 WMT News (EMNLP) dataset. The statistics of each dataset are presented in Table~\ref{table_data}.

For the SNLI dataset, considering the data distribution, we set a maximum sentence length of 15 and a vocabulary size of 11k. Each dataset represents different experimental environments, which have a critical impact on the unsupervised training of the text generation model: COCO for small-sized data with short text, SNLI for big-sized data with short text, and EMNLP for mid-sized data with long text.  

\subsection{Experimental Settings}
We implement our model using TensorFlow 1.15 and train the model with up to 200,000 iterations. For all experiments, we use the same model, loss function, and hyperparameters across the set of datasets, but different vocabulary sizes.

\subsubsection{Autoencoder}
The autoencoder is made up of an encoder GRU and a decoder GRU with 300 hidden units. We use 300-dimensional GloVe word embeddings trained on 840 billion tokens to initialize both the encoder and the decoder, and they are fine-tuned separately during training. The encoder output is normalized with l2-normalization. The input to the decoder is augmented by the output of the previous time step with a residual connection at every decoding time step. Additive Gaussian noise is injected into the encoder output and decays with a factor of 0.995 every 100 iterations. We use ADAM~\cite{kingma2014adam} optimizer with an initial learning rate of $1e^{-03}$. Gradient clipping is applied if the norm of gradients exceed 5.

\subsubsection{Generator \& Discriminators}
The code-generator and the code-discriminator are 2-layer 300-dimensional MLPs with residual connections between each layer. We use a Leaky ReLU for the activation function. The text-discriminator is a 1-layer GRU with 300 hidden units that has the same structure as the decoder. We use ADAM~\cite{kingma2014adam} optimizers and set the initial learning rate of the code-generator and two discriminators as $5e^{-06}$ and $5e^{-03}$ respectively. Gradient clipping is applied to the text-discriminator if the norm of gradients exceed 5.

\subsection{Evaluation Metrics}
The evaluation of natural language generation models is difficult since there is no single metric to measure the quality of various features of the language. In general, there are two aspects of natural language to be evaluated: quality and diversity.

\subsubsection{BLEU \& Backward BLEU}
Following previous works, we use BLEU score~\cite{papineni2002bleu} as the metric of quality. For each dataset, we first sample the same amount of generated text as the held-out test data. Then, for each generated text, the corpus-level BLEU score is calculated with the entire test data as a reference data~\cite{zhu2018texygen}. 
Reference \cite{shi2018toward} have proposed to use backward BLEU (B-BLEU) for the measurement of diversity.  For the B-BLEU score, we evaluate each of the test data given the entire generated data as a reference.
Intuitively, the BLEU score measures the precision of generated text, while the B-BLEU score measures the recall of generated text.
For both scores, a larger value indicates better performance.

\subsubsection{Fr\'echet distance}
Reference \cite{semeniuta2018accurate} proposed an automatic evaluation metric called the Fr\'echet InferSent Distance (FD), which evaluates the outputs of text generation models.
The FD calculates the Fr\'echet distance between real text and generated text in the pretrained embedding space.
This metric can capture both quality and diversity along with the global consistency of the text. Since the metric is known to be robust to the embedding model, as suggested in \cite{de2019training}, we use Universal Sentence Encoder~\cite{cer2018universal} to compute the sentence embedding of texts for our experiments.

\begin{table}[t]
\caption{Comparison of the quality and diversity of sentences generated by state-of-the-art text GANs with BLEU and B-BLEU scores on the Stanford Natural Language Inference dataset. A larger value indicates the better quality / diversity.}
\small
\centering
\begin{tabular}{C{0.18\columnwidth}|C{0.12\columnwidth}C{0.18\columnwidth}C{0.12\columnwidth}C{0.12\columnwidth}}
\toprule
Metrics & MLE & ScratchGan & ARAE* & Ours  \\ \midrule
BLEU-2  & 0.843    & 0.814   & 0.848    & \textbf{0.871} \\
BLEU-3  & 0.641    & 0.604   & 0.639    & \textbf{0.681} \\
BLEU-4  & 0.440    & 0.410   & 0.422    & \textbf{0.466} \\
BLEU-5  & 0.307    & 0.277   & 0.272    & \textbf{0.311} \\ 
\midrule
B-BLEU-2  & \textbf{0.831}    & 0.808   & 0.821 & 0.817 \\
B-BLEU-3  & 0.634    & 0.594   & 0.635 & \textbf{0.636} \\
B-BLEU-4  & 0.439    & 0.400   & 0.435 & \textbf{0.446} \\
B-BLEU-5  & 0.290    & 0.269   & 0.288 & \textbf{0.301} \\ \bottomrule
\end{tabular}
\label{table_snli_bleu_score}
\end{table}

\subsection{Experimental Results for Quality \& Diversity}
We compare our model with an MLE baseline along with other state-of-the-art text GANs, such as SeqGan~\cite{yu2017seqgan}, RankGan~\cite{lin2017adversarial}, MaliGan~\cite{che2017maximum}, and ScratchGan~\cite{de2019training} based on Texygen~\cite{zhu2018texygen}, which is an evaluation platform for text GANs. The MLE baseline is an RNN with MLE objective which has the same structure as the decoder of the proposed model. In addition, we detach the text-discriminator from our model and train the remaining part with the same training strategy as for the ARAE~\cite{zhao2018adversarially}, which is one of the most representative continuous-space text GANs.
Our RL-detached model shows superior performance over the original ARAE model (detailed information is provided in Appendix A).
We call the model ARAE* in the following sections. Every score is averaged over five runs, and they have a standard deviation smaller than 0.005.

Table~\ref{table_emnlp_bleu_bblue} reports BLEU and B-BLEU scores of text GANs trained on the EMNLP dataset.
While recently proposed ScratchGan surpasses the previous state-of-the-art text GANs by a significant margin, ConcreteGAN shows superior performance over ScratchGan. Interestingly, our implementation of an ARAE-like model (see ``ARAE*" in the table) performs better than most of the discrete-space methods. The effect of our model stands out in larger n-grams, which means that commonly-used combinations of words, such as phrases, can be generated with better quality and diversity.

Then, we compare the performance on another commonly used corpus, which is a part of the original COCO image caption dataset, and has a very small amount of training data.
As shown in Table~\ref{table_coco_bleu_bbleu}, ConcreteGAN performs better than most of the discrete-space methods in generating longer combinations of words. 
However, we find that all of the BLEU and B-BLEU scores of ARAE* are higher than those of discrete-space methods, including the proposed model. We conjecture that the lack of training data (i.e., 10k samples) cannot provide enough guidance for the RNN-based RL discriminator.

To see the effect on the dataset size, we conduct an additional experiment on the SNLI dataset, which is composed of a large amount of data with short sentences(i.e., 701k samples).
In addition to the MLE baseline, We choose ScratchGAN, ARAE*, and ConcreteGAN, which represent the discrete-space methods, continuous-space methods, and combined approaches respectively, for comparison.
Table~\ref{table_snli_bleu_score} shows the BLEU and B-BLEU scores of these three models.
With a large data for training the models, our proposed method surpasses ARAE* and achieves the best performance compared to other text GANs and the MLE baseline.

\begin{table}[t]
\caption{Comparison of the Fr\'echet distance between the real text distribution and generated text distribution in the Universal sentence embedding space. A lower value indicates the a smaller distance between the two distributions.}
\small
\centering
\begin{tabular}{C{0.15\columnwidth}C{0.14\columnwidth}C{0.18\columnwidth}C{0.14\columnwidth}C{0.14\columnwidth}}
\toprule
Dataset & MLE & ScratchGan & ARAE* & Ours \\ \midrule
COCO    & 0.245 {\scriptsize ($\pm \text{0.003}$)}   & 0.259 {\scriptsize ($\pm \text{0.005}$)}   & 0.238 {\scriptsize ($\pm \text{0.004}$)}  & \textbf{0.235} {\scriptsize ($\pm\text{0.004}$)} \\
SNLI    & 0.010 {\scriptsize ($\pm\text{0.002}$)}  & 0.011 {\scriptsize ($\pm\text{0.001}$)}  & 0.011 {\scriptsize ($\pm \text{0.001}$)} & \textbf{0.008} {\scriptsize ($\pm \text{0.001}$)} \\
EMNLP   & 0.021 {\scriptsize ($\pm \text{0.002}$)} & 0.021 {\scriptsize ($\pm \text{0.001}$)}  & 0.022 {\scriptsize ($\pm \text{0.001}$)} & \textbf{0.019} {\scriptsize ($\pm \text{0.001}$)} \\ \bottomrule
\end{tabular}
\label{table_fid}
\end{table}

\begin{table}[t]
\caption{Human evaluation score of state-of-the-art models on the EMNLP 2017 WMT News dataset. One hundred random samples generated by each model are evaluated by 10 English native workers on Amazon Mechanical Turk.}
\small
\centering
\begin{tabular}{C{0.15\columnwidth}C{0.14\columnwidth}C{0.18\columnwidth}C{0.14\columnwidth}C{0.14\columnwidth}}
\toprule
Methods & MLE & ScratchGan & ARAE* & Ours \\ \midrule
Human score    &3.157 {\scriptsize ($\pm \text{1.043}$)}   & 3.157 {\scriptsize ($\pm \text{1.048}$)}  & 3.204 {\scriptsize ($\pm \text{1.040}$)} & \textbf{3.337} {\scriptsize ($\pm \text{0.946}$)} \\ \bottomrule
\end{tabular}
\label{table_human_eval}
\end{table}

\subsection{Experimental Results for FD score}
We compare FD score between the real text distribution and the generated text distribution in the Universal Sentence Embedding space. Table~\ref{table_fid} shows the FD score of each state-of-the-art model with different learning paradigm.
Analogous to the results in the \textit{Experimental results for Quality \& Diversity Evaluation}, the FD scores of all three models on the COCO corpus are fairly high. We explain this result as a natural outcome of the lack of training data. In other corpora with large training data, our model shows the best performance, which means that it can generate text that has the most similar distribution to the real text.

\subsection{Human Evaluation}
We further conduct a human evaluation for textual sample quality of ConcreteGAN and other methods. Following previous work~\cite{nie2018relgan}, we randomly sample 100 sentences from each model and ask ten different people to score each sample on Amazon Mechanical Turk. We provide detailed criteria of human evaluation in Appendix B.
As shown in Table~\ref{table_human_eval}, the samples from ConcreteGAN are rated with the highest score compared to the state-of-the-art models of continuous-space and discrete-space models. Along with the experimental results in the previous sections, the human evaluation further demonstrate that the proposed method can generates human-like samples better than other methods.


\begin{table}[t]
\caption{The Fr\'echet distance between the latent code distribution of real and synthetic text. The lower the value is, the closer the two distributions.}
\small
\centering
\begin{tabular}{C{0.25\columnwidth}|C{0.25\columnwidth}C{0.25\columnwidth}}
\toprule
FD      & SNLI          & EMNLP  \\ \midrule
ARAE*    & 24.7          & 18.9 \\
Ours    & \textbf{15.5} & \textbf{16.2} \\ \bottomrule
\end{tabular}
\label{table_code_fd}
\end{table}


\subsection{Analyses of Code Space}
In the previous section, the textual outputs of various text GANs are compared with diversified measurements.
To demonstrate the effectiveness of the collaborative adversarial training in both the continuous and discrete spaces, we further analyze the behavior of code-generators.
As the goal of code-generator is to imitate the real distribution of text in the latent code space, we compare the outputs of the encoder and code-generator to examine the performance of code-generator.
We independently gather the encoder's outputs from the real text inputs (i.e., the test dataset) and the code-generator's outputs from the random noise inputs.

\begin{figure}[t]
\centering
\includegraphics[width=\linewidth]{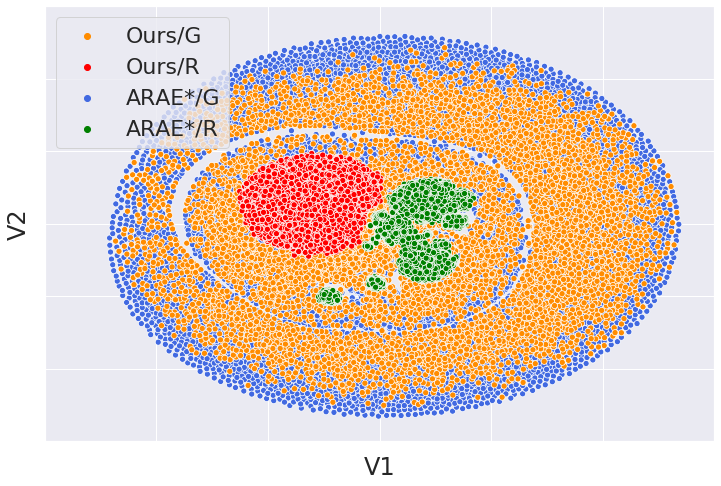}
\caption{
Comparison of latent code distribution via t-distributed stochastic neighbor embedding (t-SNE). G indicates the output distribution of the code-generator. R indicates the latent code distribution of the real text generated by the encoder. 
}
\label{fig_tSNE_code_space}
\end{figure}

\subsubsection{Analysis on t-SNE Space}
We first visualize the code distribution with t-SNE~\cite{maaten2008visualizing} for in-depth analysis.
Fig. 2 shows the t-SNE plots of four different latent code distribution; each of them represents the code-generator outputs of ConcreteGAN (Ours/G), the encoder outputs of ConcreteGAN (Ours/R), the code-generator outputs of ARAE* (ARAE*/G), and the encoder outputs of ARAE* (ARAE*/R). 
We see the encoders of both models map the real text to more compact latent spaces than the code-generators. Furthermore, the latent code space generated by our code-generator is more compact than that of ARAE* in the same embedding space. Considering that the two models (i.e., ConcreteGAN and ARAE*) employ the same encoder architecture, this result demonstrates that ConcreteGAN produces a compact and dense latent code distribution, which is more similar to the latent code space of real text.

\subsubsection{Fr\'echet Distance between Latent Code Distribution}
We further compare the Fr\'echet distance between the latent code distribution of real and synthetic text. Latent codes are obtained from the encoder and code-generator, respectively.
As the codes are represented as embedding vectors, no external model for computing the sentence-embedding is required.
As shown in Table~\ref{table_code_fd}, ConcreteGAN shows reduced Fr\'echet distance compared to ARAE* in both datasets: 24.7 to 15.15 in SNLI and 18.9 to 16.2 in EMNLP. These results demonstrate our model's superiority in generating latent codes compared to the previous baseline with a significant margin.

While the proposed model shows better performance in imitating the encoder than ARAE*, we see that the gap of Fr\'echet distance is smaller in the SNLI dataset than in the EMNLP dataset.
The average length of a sentence in EMNLP is approximately three times larger than that of SNLI, and it is more difficult for the model to encode lengthy text to a fixed-size code representation.
This observation calls for future research investigating the use of a different architecture (i.e., a CNN or Transformer) for the encoder part.

\section{Conclusion}
In this paper, we propose ConcreteGAN, a novel GAN architecture for text generation.
Unlike previous approaches, ConcreteGAN promotes the collaborative training of the continuous-space and discrete-space methods. The interplay between two adversarial trainings has a complementary effect on text generation. From a continuous-space method, our model effectively reduces the search space of RL-policy decoder. Meanwhile, discrete-space training enables the model to capture the structure of text and thereby alleviate the exposure bias, which is caused by continuous-space methods. The experimental results on three standard benchmark datasets show that ConcreteGANs outperforms state-of-the-art text GANs in terms of quality, diversity, and global consistency.

\begin{appendices}
\section{Comparison of ARAE and ARAE*}
We compare our implementation of ARAE* and the original ARAE model with SNLI dataset, since the author of the ARAE published the pretrained model. In terms of FD score, ARAE achieves 0.011 which is the same as the score of ARAE*. We further compare the BLEU and the B-BLEU score of them. Table~\ref{table_arae_arae_star} shows that our implementation of ARAE* outperforms the original ARAE with respect to sentence quality and diversity.

\begin{table}[h]
\caption{Comparison of the quality and diversity of sentences generated by ARAE and ARAE* with BLEU and B-BLEU scores on the Stanford Natural Language Inference dataset. A larger value indicates the better quality or diversity.}
\small
\centering
\begin{tabular}{C{0.18\columnwidth}|C{0.18\columnwidth}C{0.18\columnwidth}}
\toprule
Metrics & ARAE & ARAE* \\ \midrule
BLEU-2  & 0.847    & 0.848 \\
BLEU-3  & 0.620    & 0.639 \\
BLEU-4  & 0.404    & 0.422 \\
BLEU-5  & 0.263    & 0.272 \\ 
\midrule
B-BLEU-2  & 0.817    & 0.821 \\
B-BLEU-3  & 0.627    & 0.635 \\
B-BLEU-4  & 0.429    & 0.435 \\
B-BLEU-5  & 0.282    & 0.288 \\ \bottomrule
\end{tabular}
\label{table_arae_arae_star}
\end{table}

\section{Generated Samples}
We present samples generated by our proposed ConcreteGAN trained on EMNLP, COCO and SNLI dataset in Table~\ref{sample_emnlp}, Table~\ref{sample_coco} and Table~\ref{sample_snli} respectively. 

\begin{table}[h]
\caption{Randomly generated samples by ConcreteGAN trained on EMNLP dataset}
\centering
\begin{tabular}{L{0.9\columnwidth}}
    \toprule
    1. however , the number of residents affected by the emergency care will be released until april , the end of the worst record of the death . \\
    2. if they have lost their families , they will be able to get back home and hopefully we ' ll come together with them . \\
    3. there are some a little bit of interest in the coming months , it ' s unclear , " he said . \\
    4. 18 - year - old actress said : ' i try to fight my life in the end of the war , but i ' ve never done that . \\
    5. i was on the phone and i learned that i wanted to talk about you because you don ' t have any of it . \\
    6. he cannot be registered with an independent parliamentary party , but it does not have to be protected by even if they receive an act of support . \\
    7. the books , they did not deserve to be able to do something , but i don ' t think the worst . \\
    \bottomrule
\end{tabular}
\label{sample_emnlp}
\end{table}

\begin{table}[h]
\caption{Randomly generated samples by ConcreteGAN trained on COCO dataset}
\centering
\begin{tabular}{L{0.9\columnwidth}}
    \toprule
    1. red stoplights on a snow covered field . \\
    2. a plate filled with pasta and a cake on a tray . \\
    3. a cat stands on the hood of a car . \\
    4. a black and white photo of a room with a ladder in it . \\
    5. a jet plane sits on the runway while the sun rises or sets . \\
    6. a car is decorated in the dirt by a beach . \\
    7. i see into the sky beyond the light . \\

    \bottomrule
\end{tabular}
\label{sample_coco}
\end{table}

\begin{table}[h]
\caption{Randomly generated samples by ConcreteGAN trained on SNLI dataset}
\centering
\begin{tabular}{L{0.9\columnwidth}}
    \toprule
    1. a woman wearing traditional clothing is posing for a picture . \\
    2. the man is playing an instrument for a crowd of people . \\
    3. a man is giving a speech to another man at the street . \\
    4. soccer players are starting a struggle in the middle of the olympics . \\
    5. two dogs sleep in a cage . \\
    6. the man in a suit is asleep at a table . \\
    7. two people are painting pictures of a mountain . \\
    \bottomrule
\end{tabular}
\label{sample_snli}
\end{table}

\section{Human Evaluation Criteria}
The Human evaluation is based on grammatical correctness and meaningfulness and any text formatting problems (e.g., capitalization, punctuation, spelling errors, extra spaces between words and punctuations) are ignored. Workers are asked to score each sample based on the criteria shown in Table~\ref{table_human_criteria}.

\begin{table}[H]
\caption{Human evaluation criteria for 1-5 scoring.}
\centering
\begin{tabular}{L{0.23\columnwidth}|L{0.65\columnwidth}}
\toprule
Score & Criterion  \\ \midrule
5 - Excellent  & It's Grammatically correct and makes sense. \\
& For example: \textit{``if England wins the World Cup next year, it will be the most significant result the sport has seen in more than a decade.”} \\ \midrule
4 - Good  & It has some small grammatical errors and mostly make sense. \\
& For example: \textit{``it is useful to have had a doctor who forced her to release him a couple of days before she was cleared."} \\ \midrule
3 - Fair  & It has major grammatical errors but the whole still conveys some meanings. \\
& For example: \textit{``even then once again there’s a sign of that stuff is going on the way to work on Christmas eve."} \\ \midrule
2 - Poor  & It has severe grammatical errors and the whole doesn’t make sense, but some parts are still locally meaningful. \\
& For example: \textit{``we go to work for the moment in life their eyes and, i have been a different race on to go."} \\ \midrule
1 - Unacceptable  & It is basically a random collection of words \\
& For example: \textit{``i go com com com, i on on on play can go go."} \\ \bottomrule
\end{tabular}
\label{table_human_criteria}
\end{table}

\end{appendices}

\bibliographystyle{unsrt}
\bibliography{access}

\begin{IEEEbiography}[{\includegraphics[width=1in,height=1.25in,clip,keepaspectratio]{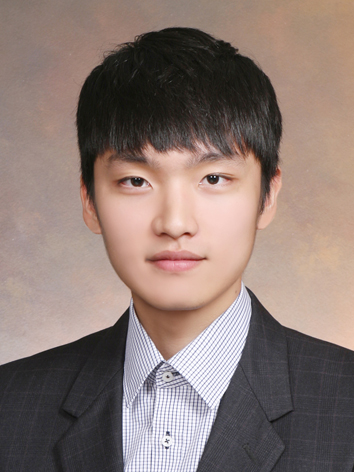}}]{Yanghoon Kim} received the B.S. degree in Automation from Tsinghua University, Beijing, China, in 2014. He is currently pursuing the Ph.D. degree in electrical and computer engineering at Seoul National University, Seoul, South Korea. His research interests include neural sequence generation and question answering systems.
\end{IEEEbiography}

\begin{IEEEbiography}[{\includegraphics[width=1in,height=1.25in,clip,keepaspectratio]{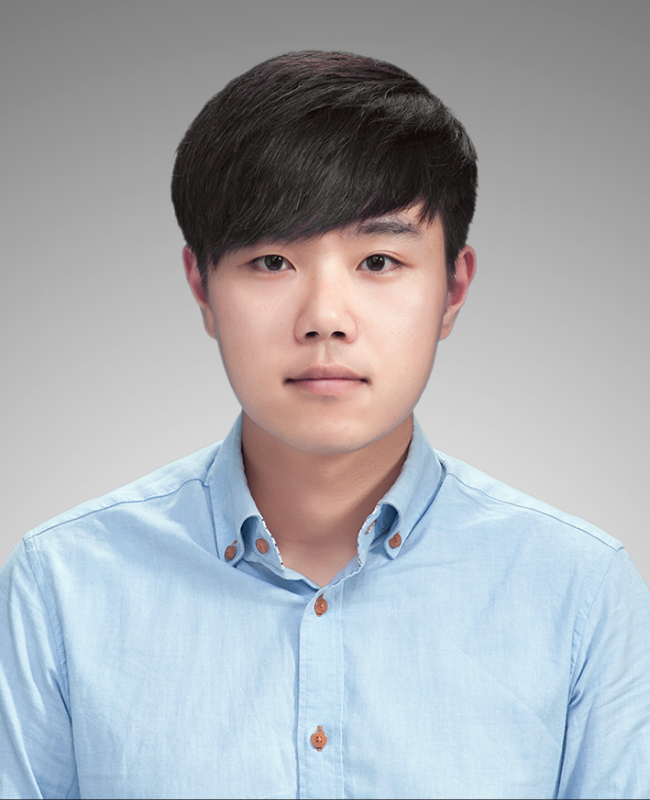}}]{Seungpil Won} received the B.S. degree in Electrical and Electronic Engineering from Yonsei University, South Korea, in 2015. He is currently pursuing the Ph.D. Degree in Electrical and Computer Engineering at Seoul National University, South Korea. His research interests focus on the areas that have benefitted from generative models, including natural language processing and computer vision.
\end{IEEEbiography}

\begin{IEEEbiography}[{\includegraphics[width=1in,height=1.25in,clip,keepaspectratio]{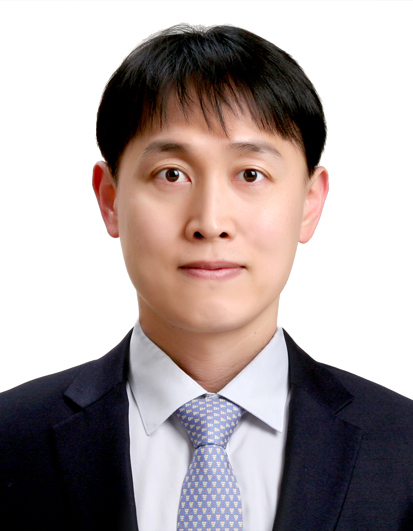}}]{Seunghyun Yoon} received the B.S. degree in Electrical and Electronics Engineering from Handong Univesity, South Korea, in 2006 and the M.S and Ph.D. degree in Electrical and Computer Engineering from Seoul National University in 2017 and 2020. He is currently a research scientist at Adobe Research, San Jose, US. His research interests are in the areas of machine learning and natural language processing (NLP). In particular, he is interested in question answering systems and learning language representation for NLP tasks.
\end{IEEEbiography}

\begin{IEEEbiography}[{\includegraphics[width=1in,height=1.25in,clip,keepaspectratio]{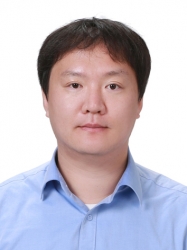}}]{Kyomin Jung} received the B.S. degree in mathematics from Seoul National University, Seoul, Korea, in 2003 
and the Ph.D. degree in Mathematics from Massachusetts Institute of Technology, Cambridge, MA, USA, in 2009.

From 2009 to 2013, he was an Assistant Professor in the department of Computer Science at KAIST. Since 2016, he has been an Associate Professor in the department of Electrical and Computer Engineering at Seoul National University (SNU). He is an adjunct professor in the department of Mathematical Sciences, SNU. His research interests include natural language processing, deep learning and applications, data analysis and web services.
\end{IEEEbiography}

\EOD

\end{document}